\documentclass[letterpaper, 10 pt, conference]{ieeeconf}  % Comment this line out if you need 

\IEEEoverridecommandlockouts                              % This command is only needed if 
\usepackage{amsmath}
\usepackage{subcaption}
\usepackage{graphicx} 
\usepackage{multirow}

\title{\LARGE \bf
Object-Centric Stereo Matching for 3D Object Detection
}

\author{Alex D. Pon and Jason Ku and Chengyao Li and Steven L. Waslander% <-this % stops a space
\thanks{Authors are with the Department of Aerospace Science and Engineering, University of Toronto. {\tt\scriptsize[alex.pon,kujason.ku,chengyao.li]@mail.utoronto.ca}, \tt\scriptsize{stevenw@utias.utoronto.ca}}% <-this % stops a space
}

\begin{document}

\maketitle
\thispagestyle{empty}
\pagestyle{empty}

%%%%%%%%%%%%%%%%%%%%%%%%%%%%%%%%%%%%%%%%%%%%%%%%%%%%%%%%%%%%%%%%%%%%%%%%%%%%%%%%
\begin{abstract}

Safe autonomous driving requires reliable 3D object detection---determining the 6 DoF pose and dimensions of objects of interest. Using stereo cameras to solve this task is a cost-effective alternative to the widely used LiDAR sensor. The current state-of-the-art for stereo 3D object detection takes the existing PSMNet stereo matching network, with no modifications, and converts the estimated disparities into a 3D point cloud, and feeds this point cloud into a LiDAR-based 3D object detector. The issue with existing stereo matching networks is that they are designed for disparity estimation, not 3D object detection; the shape and accuracy of object point clouds are not the focus. Stereo matching networks commonly suffer from inaccurate depth estimates at object boundaries, which we define as streaking, because background and foreground points are jointly estimated. Existing networks also penalize disparity instead of the estimated position of object point clouds in their loss functions. We propose a novel 2D box association and object-centric stereo matching method that only estimates the disparities of the objects of interest to address these two issues. Our method achieves state-of-the-art results on the KITTI 3D and BEV benchmarks.

\end{abstract}

%%%%%%%%%%%%%%%%%%%%%%%%%%%%%%%%%%%%%%%%%%%%%%%%%%%%%%%%%%%%%%%%%%%%%%%%%%%%%%%%
\section{Introduction}

Safe autonomous driving requires determining the six DoF pose and dimensions of objects of interest in a scene, i.e., 3D object detection. Existing methods can be categorized by the sensors they use: LiDAR~\cite{zhou2018voxelnet, yan2018second, shi2019pointrcnn, meyer2019lasernet, lang2019pointpillars}, LiDAR and camera~\cite{qi_fpointnet, ku_avod, liang2019multi}, monocular camera~\cite{ku2019monopsr, manhardt2019roi, qin2018monogrnet}, and stereo camera setups~\cite{li2019stereo_rcnn, wang2018pseudo, qin2019triangulation, chen20173d}. Methods that incorporate LiDAR measurements set the standard for 3D detection performance as LiDAR has the ability to acquire accurate depth information. However, most multi-beam LiDAR sensors remain expensive, bulky, and their returns are sparse particularly at long distances. On the other hand, acquiring depth from monocular cameras is ill-posed by nature and thus inaccurate and less reliable. Stereo camera setups are generally less expensive than LiDAR, and they resolve the under-constrained monocular problem through stereopsis. Moreover, given high resolution cameras and a large stereo baseline, stereo methods have the potential for accurate long range perception. Stereo object detection is therefore an important alternative to both monocular and LiDAR/camera methods.

Prior to our work, the state-of-the-art stereo 3D object detection method on the KITTI benchmark~\cite{geiger_kitti} was Pseudo-LiDAR~\cite{wang2018pseudo}. Pseudo-LiDAR uses the existing 3D object detector AVOD~\cite{ku_avod} and replaces the LiDAR input with a point cloud derived from the disparity output of the stereo matching network PSMNet~\cite{chang2018psmnet}. The performance loss on cars from replacing the LiDAR input is approximately 30\% AP. To understand this discrepancy, this work shows that the point clouds derived from PSMNet contain \textit{streaking} artifacts that warp the piecewise-smooth surfaces in the scene leading to significant classification and localization errors. The cause of streaking originates from the ambiguity of depth values at object edges; it can be hard to discern whether a pixel belongs to the object or the background. For such pixels, deep learning methods are often encouraged to produce depths between these two extremes~\cite{imran_depth_coefficients}. Furthermore, in deep stereo networks, closer objects are often favored during training for two main reasons. First, the inversely proportional relation between depth and disparity causes the same disparity error to have drastically different depth errors depending on the distance of objects. For example, for an object 60 m from the camera in the KITTI dataset, a disparity error of only 0.5 pixels corresponds to a large depth error of 5.1 m, but for a car 10 m away the same disparity error corresponds to a depth error of only 0.1 m. The second reason closer depths are favored during training is that there is a natural imbalance in training data. In a typical driving scene, the image is dominated by foreground pixels.

This work presents an object-centric stereo matching network, OC Stereo, to address the problems that arise from typical deep stereo matching methods. First, to resolve the streaking issue described above, we propose an object-centric representation of depth. In 3D object detection, one is primarily concerned with the objects of interest; therefore, we perform stereo matching on only object image crops and mask the ground truth background disparities during training to only penalize errors for object pixels. As a result, we avoid creating streaking artifacts in the object point clouds, and thus capture the true shapes of objects more accurately. Furthermore, as a result of only estimating disparities for the objects of interest, the runtime efficiency is significantly improved---an important aspect for safe self-driving vehicles. Second, to resolve the issue of stereo matching networks favouring closer objects, we introduce a point cloud loss that jointly penalizes the estimated position and shape of the object instances directly, and canonically resize the image crops of objects to balance the number of pixels for close and far objects.

Our main contributions are as follows: 1) A fast 2D box association algorithm that accurately matches detections between left and right images; 2) A novel object-centric stereo matching architecture that addresses the pixel imbalance problem between near and far objects and suppresses streaking artifacts in the resulting point clouds to improve 3D localization; 3) A point cloud loss within the stereo matching network to help recover object shape and to directly penalize depth errors; 4) State-of-the-art results on the KITTI 3D object detection benchmark~\cite{geiger_kitti} while running 31\% faster than the previous state-of-the-art.

%===============================================================================

\section{Related Work}
\label{sec:related_works}

\textbf{Stereo Correspondence.} Determining stereo correspondences is an active area of research in computer vision. End-to-end networks typically construct correlation layers~\cite{mayer2016large} or cost volumes~\cite{chang2018psmnet, kendall2017end, yin2019hd3, zhang2019ganet}, which can be processed efficiently on GPUs to produce high quality disparity maps. These methods already achieve less than 2\% 3-pixel error on the KITTI 2015 stereo benchmark~\cite{kitti_stereo}. However, due to the inversely proportional relation between disparity and depth, the 3-pixel error metric allows for large inaccuracies in depth especially at far distances, and thus this metric is not as meaningful for 3D object detection performance. We instead focus the stereo network on recovering meaningful object shapes and accurate depth to improve 3D detection performance.

\begin{figure*}[t]
	\begin{center}
		\includegraphics[width=0.8\linewidth]{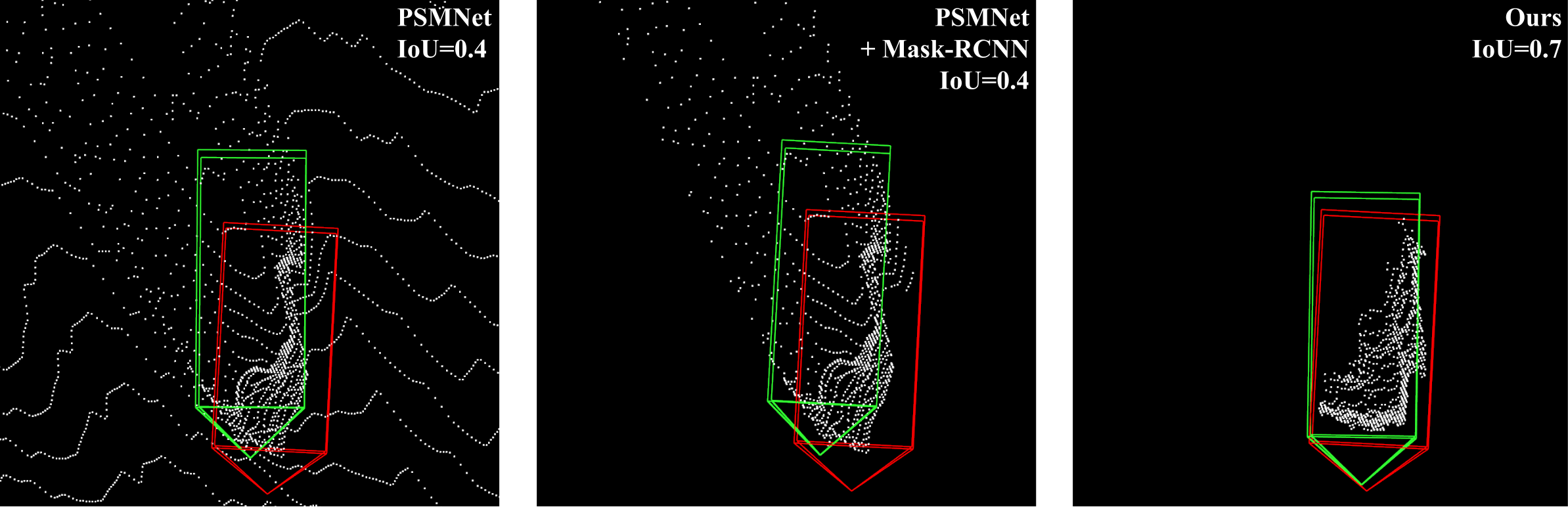}
	\end{center}
	\caption{3D localization is improved with our object-centric point cloud that avoids streaking artifacts, which occurs with PSMNet even when masked using Mask R-CNN. Ground truth and predictions are shown in \textbf{red} and \textbf{green}, respectively.}
	\label{fig:single_obj_comparison}
\end{figure*}

\textbf{Streaking Depth.} Streaking depths are a common artifact in typical stereo matching networks. Point clouds of foreground objects generated from re-projecting depth maps into 3D space are generally blurred into the background, as it can be ambiguous whether pixels belong to the object or the background. The cause of streaking has been investigated by ~\cite{imran_depth_coefficients}, who find that at ambiguous depths common loss functions prefer the mean of the foreground and background depths or do not adequately penalize estimates within this discontinuity. MonoPLiDAR~\cite{weng2019monocular} proposes to eliminate streaking artifacts with instance segmentation masks. Using a depth map estimated from a monocular image, instance segmentation masks are applied to remove background points. While their method removes some streaking, streaking still persists, as shown in Fig.~\ref{fig:single_obj_comparison}, since the instance segmentation masks are not perfect, especially at object edges where the depth ambiguities exist. Also, the full depth map is still predicted, which requires additional computation.

\textbf{Stereo 3D Object detection.} One of the early stereo 3D detectors, 3DOP~\cite{chen_3dop}, generates candidate 3D anchors which are scored and regressed to final detections using several handcrafted features. Current state-of-the-art methods are deep learning based. Pseudo-LiDAR~\cite{wang2018pseudo} adapt the 3D detectors AVOD~\cite{ku_avod} and F-PointNet~\cite{qi_fpointnet} to use point clouds from disparity maps predicted by PSMNet~\cite{chang2018psmnet}. However, this method results in point clouds with streaking artifacts, and requires additional computation by estimating depths of background areas that are not necessarily relevant for 3D object detection. On the other hand, we save computation and avoid streaking artifacts by using an object-centric approach by only estimating the depths of the objects of interest. Stereo R-CNN~\cite{li2019stereo_rcnn} creates 2D anchors that automatically associate left and right bounding boxes. These anchors are used with keypoints to estimate rough 3D bounding boxes that are later refined using photometric alignment on object image crops. TLNet~\cite{qin2019triangulation} employ 3D anchors for object correspondence and also use triangulation. However, Stereo R-CNN and TLNet perform 8\% AP and 36\% AP lower, respectively, than Pseudo-LiDAR on the KITTI moderate car category. This discrepancy suggests that explicit photometric errors and sparse anchor triangulations may be inferior to using disparity cost volumes to learn depth, and that depth estimation is the main area for improvement, which is one of the focuses of this work.

%===============================================================================

\section{Method}
\label{sec:method}

Given a pair of left and right images, $I_l$ and $I_r$, our objective is to estimate the 3D pose and dimensions of each object of interest in the scene. The main motivation behind our method is the belief that focusing on the objects of interest will result in better object detection performance. Therefore, instead of performing stereo matching on the full image, we perform stereo matching on Regions of Interest (RoIs), and only for pixels belonging to objects. This approach has three key advantages: 1) we resize the RoIs so there are a similar number of pixels for each object, which reduces class imbalance in depth values, 2) by only comparing RoIs we reduce the possible range of disparity values, and thus have faster runtime because the RoI disparity cost volumes are smaller, 3) we avoid streaking artifacts by ignoring background pixels.

Overall, the pipeline, shown in Fig.~\ref{fig:architecture}, works as follows. First, a 2D detector generates 2D boxes in $I_l$ and $I_r$. Next, a box association algorithm matches object detections across both images. Each matched detection pair is passed into the object-centric stereo network, which jointly produces a disparity map and instance segmentation mask for each object. Together, these form a disparity map containing only the objects of interest. Lastly, the disparity map is transformed into a point cloud that can be used by any LiDAR-based 3D object detection network to predict the 3D bounding boxes.

\subsection{2D Object Detector and Box Association Algorithm}

Given the stereo image pair input, we identify left and right RoIs, $l, r$, using a 2D object detector. After applying a 2D detection score threshold $t_d$, we acquire $m$ and $n$ RoIs in the left and right images, respectively. We perform association by computing the Structural SIMilarity index (SSIM)~\cite{wang2004image} for each RoI pair combination then matching the highest scores. SSIM is calculated as follows,
\begin{equation}
    SSIM(l, r)  = \frac{(2 \mu_l \mu_r + C_1)(2 \sigma_{lr} + C_2)}{(\mu_l^2 + \mu_r^2 + C_1)(\sigma_l^2 + \sigma_r^2 + C_2)},
\end{equation}
where $\mu_l, \sigma_l$, $\mu_r, \sigma_r$, are the left and right RoI pixel intensity mean and variance, $\sigma_{lr}$ is the correlation of the pixel intensities, and $C_1$ and $C_2$ are constants to prevent division by zero. This metric is calculated per image channel and averaged. Our assumption is that objects in the left and right images have similar appearance as SSIM measures the visual similarity between two images emphasizing relations of spatially close pixels.

Each RoI is then interpolated to a standard size. The SSIM index is calculated between each left and right RoI. The algorithm determines association by going in order of highest to lowest scoring SSIM indices using the image with fewer boxes. Once a box is associated, it is removed for faster comparison. At the end of the algorithm, unmatched boxes are considered false positives and removed.

\begin{figure*}[t!]
	\begin{center}
		\includegraphics[width=0.81\linewidth]{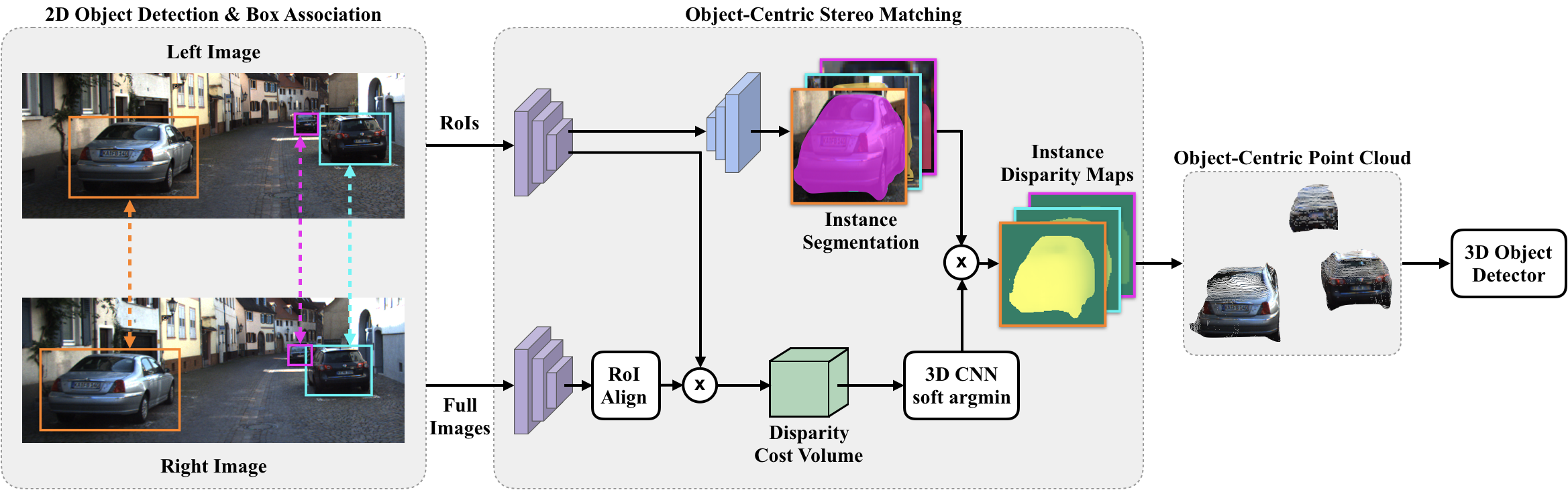}
	\end{center}
	\caption{A 2D detector and box association algorithm determine associated RoIs. Our stereo matching network estimates disparities with a 3D CNN and soft argmin operation~\cite{chang2018psmnet} for object pixels using the RoIs and instance segmentation. These are converted to a 3D point cloud and can be inputted to any LiDAR-based 3D object detector. X indicates multiplication.}
	\label{fig:architecture}
\end{figure*}

To improve the robustness of the association, we ensure that the difference between associated 2D bounding box centres are within an adaptive box center threshold. MonoPSR~\cite{ku2019monopsr} shows the depth of objects is well correlated with bounding box height. This means that closer objects should have larger disparities while further objects should have smaller disparities. Using the KITTI dataset, we model the relationship between box height and centre disparity using linear regression. Based on an RoI's box height, the data provides the expected centre disparity for its associated box. We therefore constrain the maximum distance between box centers of associated RoIs to be within three standard deviations of the expected disparity. Boxes that do not satisfy these conditions are ignored for the SSIM calculation, further improving the speed and accuracy of the associations. An example of the corresponding RoIs is shown in Fig.~\ref{fig:architecture}.

While our method is reliant on 2D detection quality, we believe using a 2D detector is actually advantageous because 2D detection is a mature field with robust performance. Radosavovic et al.~\cite{data_distillation_radosavovic} even claim that the current performance of 2D detectors is accurate enough that detectors can be trained using data it inferences---self-training. In Sec.~\ref{sec:result} we show our 2D detections have higher AP compared to other state-of-the-art.

\subsection{Object-Centric Stereo Matching}

Given the associated RoIs, we perform stereo matching to estimate a canonically resized disparity map of dimensions $w \times h$ per object. Within the RoIs, disparities are learned only for pixels belonging to the object to remove depth ambiguity and thus depth streaking artifacts.

\textbf{Local Disparity Formulation.} We estimate the horizontal pixel shift, or disparity, within the aligned left RoI and right RoI. We refer to this disparity estimation as \textit{local} compared to the \textit{global} disparity shift between the pixels of the full-sized left right stereo pair. This local formulation leads to positive and negative ground truth disparities. To obtain the ground truth local disparities we first start by forming an array of the local RoI images coordinates $i_l$ of the left RoI,
\begin{equation}
    i_l = 
        \begin{bmatrix}
        0 & \dots & w \\
        \vdots  & \ddots & \\
        0 & \dots & w 
        \end{bmatrix}.
\end{equation}
The global horizontal image coordinates $x_l$ of the left RoI is 
\begin{equation}
    x_l = i_l + e_l,
\end{equation}
where $e_l$ is the horizontal coordinate of the left RoI's left edge. We calculate the disparity map corresponding to the resized RoIs by performing a nearest neighbor resizing of the ground truth global disparity map $d_g$ to the canonical size, $w \times h$. Therefore, the corresponding right global image coordinates are calculated as,
\begin{equation}
    x_r = x_l - d_g.
\end{equation}
These coordinates are normalized to the local coordinate system,
\begin{equation}
    i_r = \frac{(x_r - e_r)}{w_b}  w, 
\end{equation}
where $e_r$ is the horizontal coordinate of the right RoI's left edge and $w_b$ is the width of the non-resized RoI bounding box. Lastly, the local disparity of the crops can be calculated as
\begin{equation}
    d_l = i_l - i_r.
\end{equation}
During training, we use ground truth instance segmentation masks to only train on disparity values corresponding to the object. This formulation removes depth ambiguity at edges and removes streaking artifacts as shown in Fig.~\ref{fig:single_obj_comparison}. 

During inference, we mask background pixels using a predicted instance segmentation mask. From the predicted local disparity map $d_l^*$ we calculate the global disparity map $d_g^*$ by reversing the above steps. The corresponding depth map $D$ is calculated from the known horizontal focal length $f_u$ and baseline $b$ as,
\begin{equation}
    D = \frac{f_u b}{d_g^*}.
\end{equation}
Lastly, each pixel $(u,v)$ of this depth map is converted into a 3D point as
\begin{equation}
    x = \frac{(u - c_u)z}{f_u},
    y = \frac{(v - c_v)z}{f_v},
    z = D(u,v),
\end{equation}
where $(c_u, c_v)$ is the camera center and $f_v$ is the vertical focal length.

\textbf{Object-Centric Stereo Architecture.} The described object-centric stereo depth formulation is flexible with most stereo depth estimation networks. In our implementation, we build on PSMNet~\cite{chang2018psmnet} with key modifications.

We use the same feature extractor as \cite{chang2018psmnet}, but use one for the RoIs and another for the full-sized images. Despite only comparing RoIs, we leverage global context by performing RoI Align~\cite{he2017mask} on the full-sized image feature extractor outputs. The resulting features from the left image are multiplied with the left crop feature map, and the features from the right image are multiplied with the right crop feature map. To estimate disparity, the left and right feature maps are concatenated to form a 4D disparity cost volume (height $\times$ width $\times$ disparity range $\times$ feature size) as in \cite{chang2018psmnet}. Importantly, however, our input size and disparity range are smaller than what would be used for global disparity estimation because the local disparity range between two RoIs is smaller than the global disparity range between two full-sized images. As a result, we create a set of smaller cost volumes, which results in a faster runtime.

To predict the instance segmentation map, only the left feature maps are used. The instance segmentation network consists of a simple decoder; the feature map is processed by three repeating bilinear up-sampling and $3 \times 3$ convolutional layers resulting in a $w \times h$ instance segmentation mask. For each instance, the predicted segmentation mask is applied to the estimated local disparity map. To deal with overlapping instance masks, each local disparity is converted to a global disparity, resized to the original box size, and placed in farthest to closest depth order in the scene.

\textbf{Point Cloud Loss.} Similar to \cite{chang2018psmnet}, we use the smooth L1 loss to compare the predicted local disparity $d_l^*$ and the ground truth local disparity $d_l$. Penalizing the disparities directly, however, is non-ideal because it places less emphasis on far objects due to the inverse relation between disparity and depth. For example, for a car 60 m from the camera in the KITTI dataset, a disparity error of only 0.5 pixels corresponds to a large depth error of 5 meters, but for a car 10 m away the same disparity error corresponds to a depth error of only 0.13 m. An unwanted consequence of computing loss from disparity estimates is that drastically different depth errors can have the same loss value.

Therefore, we transform the predicted disparities to a point cloud. We then use the smooth L1 loss to compare each object's point cloud $p_c$ with its ground truth point cloud $p_g$. Since we are concerned about 3D localization, this loss is more suitable as it directly penalizes predicted 3D point positions and resolves the lack of emphasis on far depths described above.

\subsection{3D Box Regression Network}
One of the benefits of our pipeline is that we can use the estimated point cloud as input to any 3D object detector that processes point clouds. In our implementation we build on the AVOD~\cite{ku_avod} architecture and make two modifications. We first note that in AVOD, the RoI cropping operation of the second stage returns identical BEV features regardless of the vertical location of a regressed anchor, or proposal. As well, since our stereo point cloud does not contain ground points, we append the proposal's 3D position information to the feature vector used to regress each 3D proposal. We also check if the final 3D bounding boxes align with the 2D detections in the first stage. If a 3D box projected into the image plane does not overlap with a 2D detection by at least 0.5 IoU, it is removed.
%===============================================================================

\begin{figure}[t!]
	\begin{center}
		\includegraphics[width=0.75\linewidth]{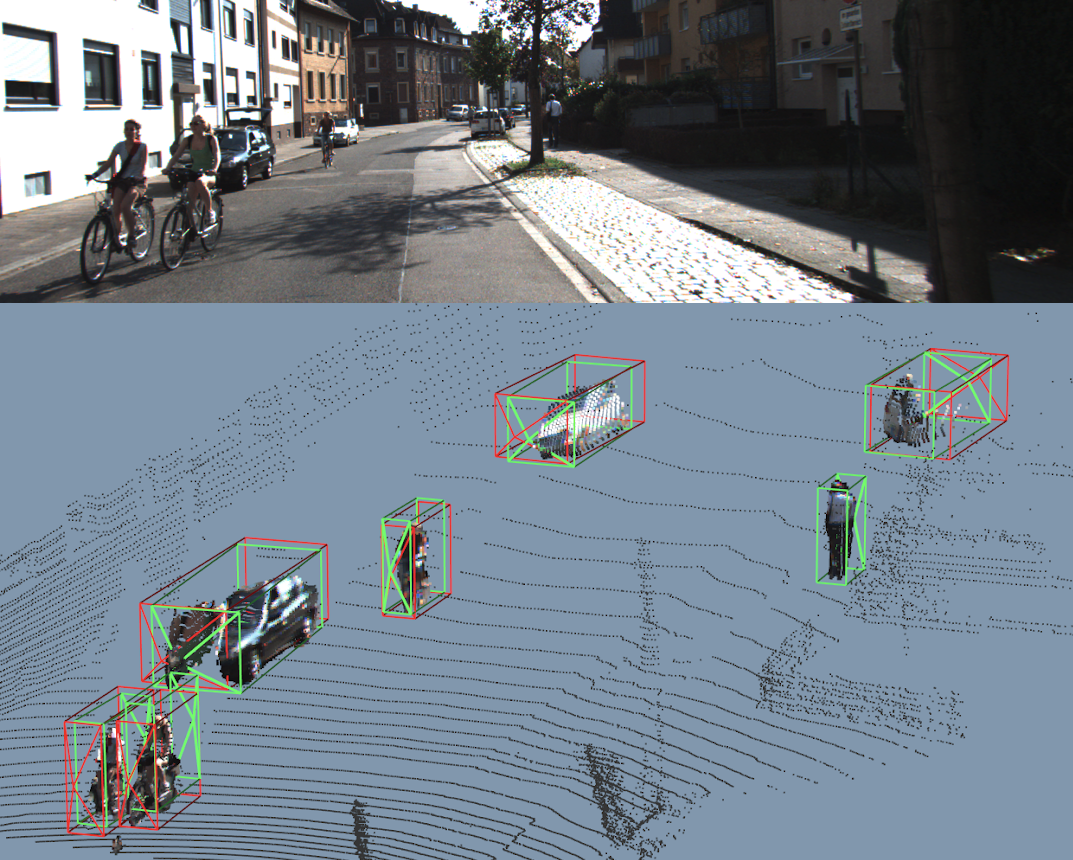}
	\end{center}
	\caption{Qualitative results on KITTI. Ground truth and predictions are in \textbf{red} and \textbf{green}, respectively. Colored points are predicted by our stereo matching network while LiDAR points are shown in black for visualization purposes only.}
	\label{fig:detections}
\end{figure}

\section{Implementation}
\label{sec:implementation}

\textbf{2D Detector and Box Association.} We use MS-CNN~\cite{cai_mscnn} as our 2D detector because it has fast runtime speed and high accuracy. The RoIs are cropped and bilinearly resized to $128 \times 128$ for association and $224 \times 224$ for local stereo matching.

\textbf{Object-Centric Stereo Network.} During stereo matching, the minimum and maximum local disparities are set as -64 and 90 pixels. This range was found by calculating the range of local disparities for randomly jittered ground truth 2D boxes that maintain a minimum 0.7 IoU with the original box. For faster convergence, the feature extractors are pre-trained on full depth maps from the SceneFlow dataset~\cite{mayer2016large} and depth completed LiDAR scans from the training split of the KITTI object detection dataset. No training is done on the KITTI raw or stereo datasets because these datasets contain overlapping samples with the object detection dataset. The object-centric stereo network, which leverages these feature extractors, is fine-tuned on crops of depth completed LiDAR scans. Depth completion is used for additional training points and faster convergence, and to remove the erroneous depths due to the differing locations of the camera and LiDAR sensor~\cite{amiri2019semisup}. The depth completion method used is \cite{ku2018defense} because it is structure preserving and does not contain streaking artifacts. The ground truth instance segmentation masks used to mask the background disparity are created by projecting the points within the ground truth 3D boxes into the image. These instance masks exactly correspond to the pixels belonging to the object, but they are not smooth due to the depth completion, so the instance segmentation network is instead trained using masks from \cite{ChenCVPR14}. For optimization, Adam was used with a batch size of 16 and a learning rate of 0.001 for 8 epochs then 0.0001 for 4 more epochs.

\textbf{3D Object Detection.} For the 3D object detector, we use AVOD~\cite{ku_avod} to compare with Pseudo-LiDAR. With a batch size of 1, the Adam optimizer is used with a learning rate of 0.0001 for 50000 steps then decayed to 0.00001 and stopped using early stopping. The data augmentation used was horizontal flipping and PCA jittering.

\begin{table*}[t!]
	\small
	\centering
	\tabcolsep=0.11cm
	\begin{tabular}{|c||c@{~/~}cc@{~/~}cc@{~/~}c||c@{~/~}cc@{~/~}cc@{~/~}c|}
		\hline
		\multirow{2}{*}{Method} & \multicolumn{6}{c||}{0.5 IoU} & \multicolumn{6}{c|}{0.7 IoU} \\
		\cline{2-13} &
		\multicolumn{2}{c}{Easy} & \multicolumn{2}{c}{Moderate} & \multicolumn{2}{c||}{Hard} &
		\multicolumn{2}{c}{Easy} & \multicolumn{2}{c}{Moderate} & \multicolumn{2}{c|}{Hard} \\ \hline
		TLNet~\cite{qin2019triangulation} & 62.46 & 59.51 & 45.99 & 43.71 & 41.92 & 37.99 & 29.22 & 18.15 & 21.88 & 14.26 & 18.83 & 13.72 \\
		\hline
		S-RCNN~\cite{li2019stereo_rcnn} & 87.13 & 85.84 & 74.11 & 66.28 & 58.93 & 57.24 & 68.50 & 54.11 & 48.30 & 36.69 & 41.47 & 31.07 \\
		\hline
		PL-FP~\cite{wang2018pseudo} & 89.8 & 89.5 & 77.6 & 75.5 & 68.2 & 66.3 & 72.8 & 59.4 & 51.8 & 39.8 & 44.0 & 33.5 \\
		\hline
		PL-AVOD~\cite{wang2018pseudo} & 89.0 & 88.5 & 77.5 & 76.4 & 68.7 & 61.2 & 74.9 & 61.9 & 56.8 & 45.3 & 49.0 & 39.0 \\
		\hline
		Ours & \textbf{90.01} & \textbf{89.65} & \textbf{80.63} & \textbf{80.03} & \textbf{71.06} & \textbf{70.34} & \textbf{77.66}  & \textbf{64.07} & \textbf{65.95} & \textbf{48.34} & \textbf{51.20} & \textbf{40.39} \\
		\hline
	\end{tabular}
	\caption{\textbf{Car Localization and Detection.} $AP_{BEV}$ / $AP_{3D}$ on \textit{val}.}
	\label{tab:kitti_val_cars}
\end{table*}

\begin{table*}[t!]
	\small
	\centering
% 	\addtolength{\tabcolsep}{-1pt}
	\tabcolsep=0.13cm{
    % \scalebox{0.95}{
	\begin{tabular}{|c||c@{~/~}cc@{~/~}cc@{~/~}c||c@{~/~}cc@{~/~}cc@{~/~}c|}
		\hline
		\multirow{2}{*}{Method}  & \multicolumn{6}{c||}{Pedestrians} & \multicolumn{6}{c|}{Cyclists} \\
		 &
		\multicolumn{2}{c}{Easy} & \multicolumn{2}{c}{Moderate} & \multicolumn{2}{c||}{Hard} &
		\multicolumn{2}{c}{Easy} & \multicolumn{2}{c}{Moderate} & \multicolumn{2}{c|}{Hard} \\ \hline
		PSMNet + AVOD & 36.68 & 27.39 & 30.08 & 26.00 & 23.76 & 20.72 & 36.12 & 35.88 & 22.99 & 22.78 & 22.11 & 21.94 \\
		PL-FP~\cite{wang2018pseudo} & 41.3 & 33.8 & 34.9 & 27.4 & 30.1 & 24.0 & 47.6 & 41.3 & \textbf{29.9} & 25.2 & \textbf{27.0} & \textbf{24.9} \\
		\hline
		Ours  & \textbf{44.00} & \textbf{34.80} & \textbf{37.20} & \textbf{29.05} & \textbf{30.39} & \textbf{28.06} & \textbf{48.20} & \textbf{45.59} & 27.90 & \textbf{25.93} & \textbf{26.96} & 24.62  \\
		\hline
	\end{tabular}
	}
	\caption{\textbf{Pedestrian and Cyclist Localization and Detection.} $AP_{BEV}$ / $AP_{3D}$ on \textit{val}. We note that \cite{wang2018pseudo} only provides values up to one decimal place.}
	\label{tab:kitti_val_ped_cyc}

\end{table*}

\begin{table*}[t!]
	\small
	\centering
	
	\begin{tabular}{|c||ccc||ccc|}
		\hline
		\multirow{2}{*}{Method} & \multicolumn{3}{c||}{BEV AP} & \multicolumn{3}{c|}{3D AP} \\
		            &    Easy & Moderate &  Hard &  Easy & Moderate &  Hard \\ \hline
		S-RCNN~\cite{li2019stereo_rcnn} & 61.67 & 43.87 & 36.44 &  49.23 & 34.05 & 28.39 \\ \hline
		PL-FP~\cite{wang2018pseudo}   & 55.0 & 38.7 & 32.9 & 39.7 & 26.7 & 22.3 \\
		PL-AVOD~\cite{wang2018pseudo} & 66.83 & 47.20 & 40.30 &  \textbf{55.40} & 37.17 & 31.37 \\
		\hline
		Ours   & \textbf{66.97} & \textbf{54.16} & \textbf{46.70} & 55.11 & \textbf{38.80} & \textbf{31.86} \\ \hline
	\end{tabular}
	\caption{\textbf{Car Localization and Detection.} \emph{$AP_{BEV}$} and \emph{$AP_{3D}$} on KITTI \emph{test}.}
	\label{tab:kitti_test_cars}
\end{table*}

\begin{table*}[t!]
	\small
	\centering
	\tabcolsep=0.13cm
	\begin{tabular}{|c||c@{~/~}cc@{~/~}cc@{~/~}c||c@{~/~}cc@{~/~}cc@{~/~}c|}
		\hline
		\multirow{2}{*}{Method}  & \multicolumn{6}{c||}{Pedestrians} & \multicolumn{6}{c|}{Cyclists} \\
		&
		\multicolumn{2}{c}{Easy} & \multicolumn{2}{c}{Moderate} & \multicolumn{2}{c||}{Hard} &
		\multicolumn{2}{c}{Easy} & \multicolumn{2}{c}{Moderate} & \multicolumn{2}{c|}{Hard} \\ \hline
		PL-FP~\cite{wang2018pseudo} & 31.3 & \textbf{29.8} & \textbf{24.0} & \textbf{22.1} & 21.9 & 18.8 & 4.1 & 3.7 & 3.1 & 2.8 & 2.8 & 2.1 \\
		PL-AVOD~\cite{wang2018pseudo} & 27.5 & 25.2 & 20.6 & 19.0 & 19.4 & 15.3 & 13.5 & 13.3 & 9.1 & 9.1 & 9.1 & 9.1 \\
		\hline
		Ours  & \textbf{35.12} & 28.14 & 23.23 & 21.85 & \textbf{22.56} & \textbf{20.92} & \textbf{34.77} & \textbf{32.66} & \textbf{22.26} & \textbf{21.25} & \textbf{21.36} & \textbf{19.77}  \\
		\hline
	\end{tabular}
	\caption{\textbf{Pedestrians and Cyclists Localization and Detection.} \emph{$AP_{BEV}$} and \emph{$AP_{3D}$} on KITTI \textit{test}.}
	\label{tab:kitti_test_ped_cyc}
\end{table*}

\begin{table*}[t!]
	\small
	\centering
	\tabcolsep=0.13cm
	\begin{tabular}{|c||ccc||ccc||ccc|}
		\hline
		\multirow{2}{*}{Metric}  & \multicolumn{3}{c||}{Left} & \multicolumn{3}{c||}{Right} & \multicolumn{3}{c|}{Stereo} \\
		& Easy & Moderate & Hard & Easy & Moderate & Hard & Easy & Moderate & Hard \\
		\hline
		S-RCNN~\cite{li2019stereo_rcnn} & 98.73 & 88.48 & 71.26 & 98.71 & 88.50 & 71.28 & \textbf{98.53} & 88.27 & 71.14 \\ \hline
		Ours     & 97.77 & 89.93 & 80.53 & 98.23 & 90.09 & 80.50 & 97.13 & 89.63 & 80.02 \\
		Ours Adaptive Thresh   & \textbf{98.87} & \textbf{90.53} & \textbf{81.05} & \textbf{98.92} & \textbf{90.50} & \textbf{80.88} & 98.44 & \textbf{90.38} & \textbf{80.71} \\ \hline
	\end{tabular}
	\caption{\textbf{Stereo 2D AP.} 2D detections and stereo box correspondence AP on \textit{val}.}
	\label{tab:kitti_2d_AP}
\end{table*}

\begin{table}[t!]
\centering
    \begin{tabular}{|l|c|}
		\hline
		\multicolumn{1}{|c|}{Version} & $AP_{BEV}$ \\
		\hline
		Baseline~\cite{wang2018pseudo} & 56.8  \\
		Baseline + Pre-trained weights &  57.10 \\
		Baseline + Mask-RCNN~\cite{he2017mask}   &  49.20 \\
		Local   &  64.90 \\
		Local + AVOD mods.  &  65.40 \\
		Local + AVOD mods. + PC Loss & \textbf{65.95} \\
		\hline
	\end{tabular}
   \caption{\textbf{Ablation Studies}. Comparisons of $AP_{BEV}$ at 0.7 IoU using \cite{wang2018pseudo} as the baseline. Local is our object-centric stereo network.}
   \label{tab:ablation}
\end{table}

\begin{table}[t!]
\centering
   \begin{tabular}{|l|c|}
		\hline
		\multicolumn{1}{|c|}{Stage} & Runtime (s) \\
		\hline
		MS-CNN~\cite{cai_mscnn}   & 0.080 \\
		Box Association & 0.009 \\ 
		Stereo Matching & 0.161  \\
		AVOD~\cite{ku_avod}  & 0.100 \\
		\hline
		Total & \textbf{0.350} \\
		\hline
	\end{tabular}
   \caption{\textbf{Runtime Analysis}. Runtime for each stage of our method. Our total runtime is faster than the previous state-of-the-art: PSMNet + AVOD (0.410s + 0.100s).}
   \label{tab:runtime}
\end{table}

\section{Experimental Results}
\label{sec:result}

We compare against the state-of-the-art, perform ablation studies, and provide qualitative results (Fig.~\ref{fig:detections}) using the KITTI dataset~\cite{geiger_kitti}. The KITTI dataset contains 7481 training images and 7518 test images, and categorizes objects in three categories: Easy, Moderate, and Hard based on 2D box height, occlusion, and truncation. To compare with the state-of-the-art, we follow the 1:1 training to validation split of \cite{chen_mv3d, ku_avod, qi_fpointnet} and the standard practice of comparing BEV and 3D AP performance using IoUs of 0.5 and 0.7. We also benchmark our results on the online KITTI test server.

\subsection{3D AP Comparison with the State-of-the-Art}
As mentioned in Sec.~\ref{sec:related_works} 3-pixel error is not indicative of 3D object detection performance as it allows large inaccuracies in depth. An alternative is comparing depth map errors. The depth map RMSE for our method and Pseudo-LiDAR is 1.60 m and 1.81 m, respectively, when comparing the same pixels that are predicted in both Pseudo-LiDAR and our depth maps. However, we believe object detection AP is more meaningful than depth map metrics because depth map errors are not as indicative of the shape of each object. We therefore use object detection AP for the remaining comparisons.

We compare to the state-of-the-art using $AP_{BEV}$ and $AP_{3D}$ on the validation set in 
Tab.~\ref{tab:kitti_val_cars} and Tab.~\ref{tab:kitti_val_ped_cyc}. For the car class, we outperform the state-of-the-art in all categories. Most noticeably, we have a 9.2\% AP increase in the BEV moderate category at 0.7 IoU, which is used to rank methods on the KITTI online server. For pedestrians and cyclists we surpass Pseudo-LiDAR with F-PointNet (PL-FP) in all but three categories and tie up to rounding error on hard cyclist BEV. We also surpass the performance of Pseudo-LiDAR implemented with AVOD, as shown in the top row, which indicates that much of the performance improvement for PL-FP can be attributed to F-PointNet. We leave using our stereo outputs on different 3D object detectors as future work. Results on the test set show similar performance improvements for our method in Tab.~\ref{tab:kitti_test_cars} and Tab.~\ref{tab:kitti_test_ped_cyc}.

In Tab.~\ref{tab:runtime} we provide runtime analysis using a Titan Xp GPU. Our method runs faster than the current state-of-the-art, Pseudo-LiDAR, by 160 ms. They run PSMNet (0.410s) and AVOD (0.100s), while our entire pipeline takes 0.350s. Our speed boosts can be attributed to the fact we only estimate disparities for RoIs, and our object-centric formulation builds a set of smaller disparity cost volumes.

\subsection{2D AP Comparison with Box Association}

We compare our box association method with Stereo-RCNN~\cite{li2019stereo_rcnn} using 2D and stereo AP. Stereo AP~\cite{li2019stereo_rcnn} is calculated by requiring a minimum 0.7 IoU with the ground truth box for the left and right bounding boxes and for the left and right bounding boxes to belong to the same object. As shown in Tab.~\ref{tab:kitti_2d_AP} the 2D detector MS-CNN and our box association algorithm outperforms or has comparable results to Stereo R-CNN. In particular, there is a 9.57\% AP improvement in the hard category. Moreover, in Tab.~\ref{tab:kitti_2d_AP}, there is a minimal decrease from our left and right AP to our stereo AP, which demonstrates that minimal performance is lost by performing association.

\subsection{Effect of Local Stereo Depth Estimation}
In Tab.~\ref{tab:ablation} we provide ablation studies. The baseline used is Pseudo-LiDAR~\cite{wang2018pseudo}. The third row of the table shows that we outperform an additional baseline that only keeps foreground depth pixels from PSMNet using a version of Mask R-CNN~\cite{he2017mask}. As shown in Fig.~\ref{fig:single_obj_comparison}, this is in part because this Mask-RCNN baseline still contains ambiguous depths and is susceptible to streaking artifacts. We note that our object-centric disparity formulation makes our method robust to some erroneous segmentation predictions because our network is trained with only object pixels, so it learns to set some background pixels to the object depth to help maintain object shape. Tab.~\ref{tab:ablation} shows the benefits of our method (Local), pre-training AVOD on depth completed LiDAR, appending anchor information to AVOD's proposal regression, and employing our point cloud loss.

\bibliographystyle{IEEEtran}
\bibliography{IEEEtran}

\end{document}